\documentclass{article} 
\usepackage{iclr2024_conference,times}
\usepackage{graphicx}

\usepackage{amsmath,amsfonts,bm}

\def\eqref#1{equation~\ref{#1}}

\def\1{\bm{1}}

\DeclareMathAlphabet{\mathsfit}{\encodingdefault}{\sfdefault}{m}{sl}
\SetMathAlphabet{\mathsfit}{bold}{\encodingdefault}{\sfdefault}{bx}{n}

\usepackage{url}
\usepackage{xcolor}
\usepackage[linkbordercolor=teal,citebordercolor=teal]{hyperref}

\title{hyper-parameter tuning for text guided image editing}

\author{Shiwen Zhang\\
\\
\texttt{witcherofresearch@gmail.com} \\
}

\iclrfinalcopy 
\begin{document}

\maketitle

\begin{abstract}
The test-time finetuning text-guided image editing method,  Forgedit, is capable of tackling general and complex image editing problems given only the input image itself and the target text prompt. During finetuning stage,  using the same set of finetuning hyper-paramters every time for every given image, Forgedit remembers and understands the input image in 30 seconds. During editing stage, the workflow of Forgedit might seem complicated. However, in fact, the editing process of Forgedit is not more complex than previous SOTA Imagic, yet completely solves the overfitting problem of Imagic. In this paper, we will elaborate the workflow of Forgedit editing stage with examples. We will show how to tune the hyper-parameters in an efficient way to obtain ideal editing results.

\end{abstract}

Forgedit \cite{zhang2023forgedit} is a new SOTA text-guided image editing method with implementation open-sourced \citep{Zhang_Pytorch_implementation_of_2024}. It is designed to be a very general editing method, capable of conducting almost every kind of text-guided editing, including yet not limited to changing actions, adding or replacing or removing objects, changing backgrounds, changing textures and styles, changing facial expressions etc. Compared with other image editing methods, Forgedit demonstrates superior  generalization ability and is especially capable of performing precise non-rigid editing on real images, even implemented with the outdated Stable Diffusion 1.4 \citep{Rombach2021HighResolutionIS}, surpassing previous SOTA Imagic \citep{DBLP:conf/cvpr/KawarZLTCDMI23} implemented with Imagen\citep{DBLP:conf/nips/SahariaCSLWDGLA22}.

During the finetuning stage, Forgedit utilizes a set of pre-defined universal hyper-parameters to memorize and understand the input image, which means that no matter what the input image is, the finetuning process will only be executed once. Such one-time finetuning without further adjustment on hyper-parameters ensures the efficiency of Forgedit. Forgedit relies on joint vision-language optimization to achieve  fast convergence speed, resulting in 30 seconds fintuning time on one nvidia-a100 GPU, much faster than previous SOTA Imagic \citep{DBLP:conf/cvpr/KawarZLTCDMI23}. However, although Forgedit utilize BLIP \citep{Li2022BLIPBL} to generate source prompt for optimization to ease overfitting, it still suffers from overfitting issues for some editing cases. In order to tackle these hard cases, during editing stage, Forgedit introduces forgetting strategies based on our novel observation of UNet's disentangled property, i.e., UNet encoder learns space and structure, UNet decoder learns appearance and texture. Such utilization of UNet disentanglement solves the overfitting issue in most cases.

In discriminative deep learning, for example, image and video classification, most methods \citep{he2016deep,zhangv4d,zhang2022tfcnet, tong2022videomae,huang20204d,huang20214d,wang2016temporal,wang2021tdn} are implicit learning methods, which means that the algorithm learns the label directly from the video without utilizing external supervision signals. Explicit learning methods, for example, KINet \citep{zhang2020knowledge}, introduces auxiliary external pseudo labels  to strengthen the implict learning process. For image editing methods, which are solved by generative models, many methods tend to conduct the image editing in explict ways by asking users to provide extra conditions \citep{zhang2023adding} or masks \citep{Rombach2021HighResolutionIS}. In contrast, our Forgedit is an implict method which learns everything directly from the original image and source text prompt auto-generated by BLIP \citep{Li2022BLIPBL}, which actually means that for the learning process, the only exteranl input is the original image, everything else is obtained by the system itself implicitly.

\begin{figure*}[!ht]
  \centering
   \includegraphics[width=1\linewidth]{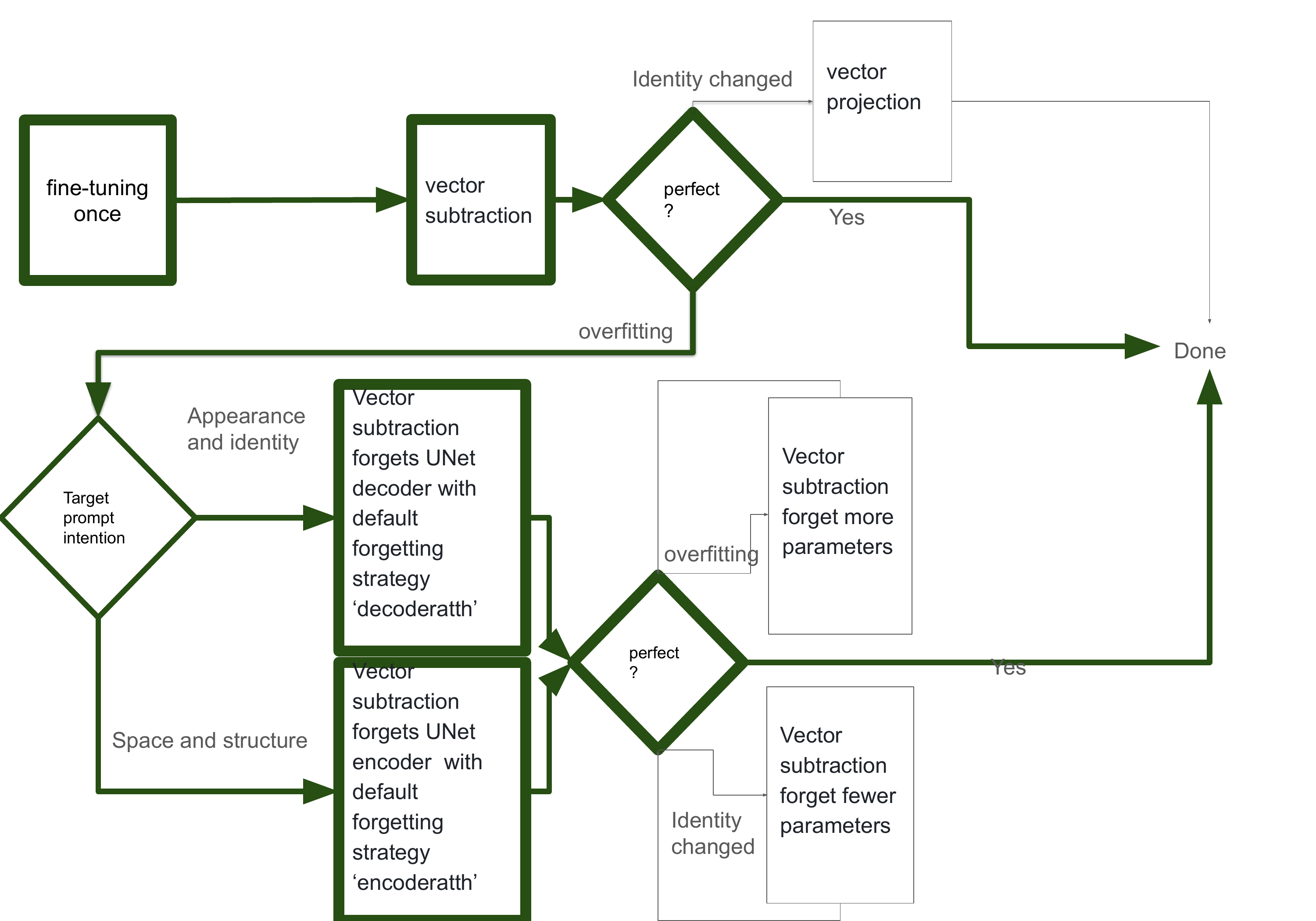}

   \caption{The workflow of Forgedit, the most usual flow of editing process is highlighted in the figure, i.e. simple vector subtraction and default forgetting strategies according to our findings of the disentangle rules of UNet.}
   \label{hyperworkflow}
\end{figure*}

\begin{figure*}[!ht]
  \centering
   \includegraphics[width=1.01\linewidth]{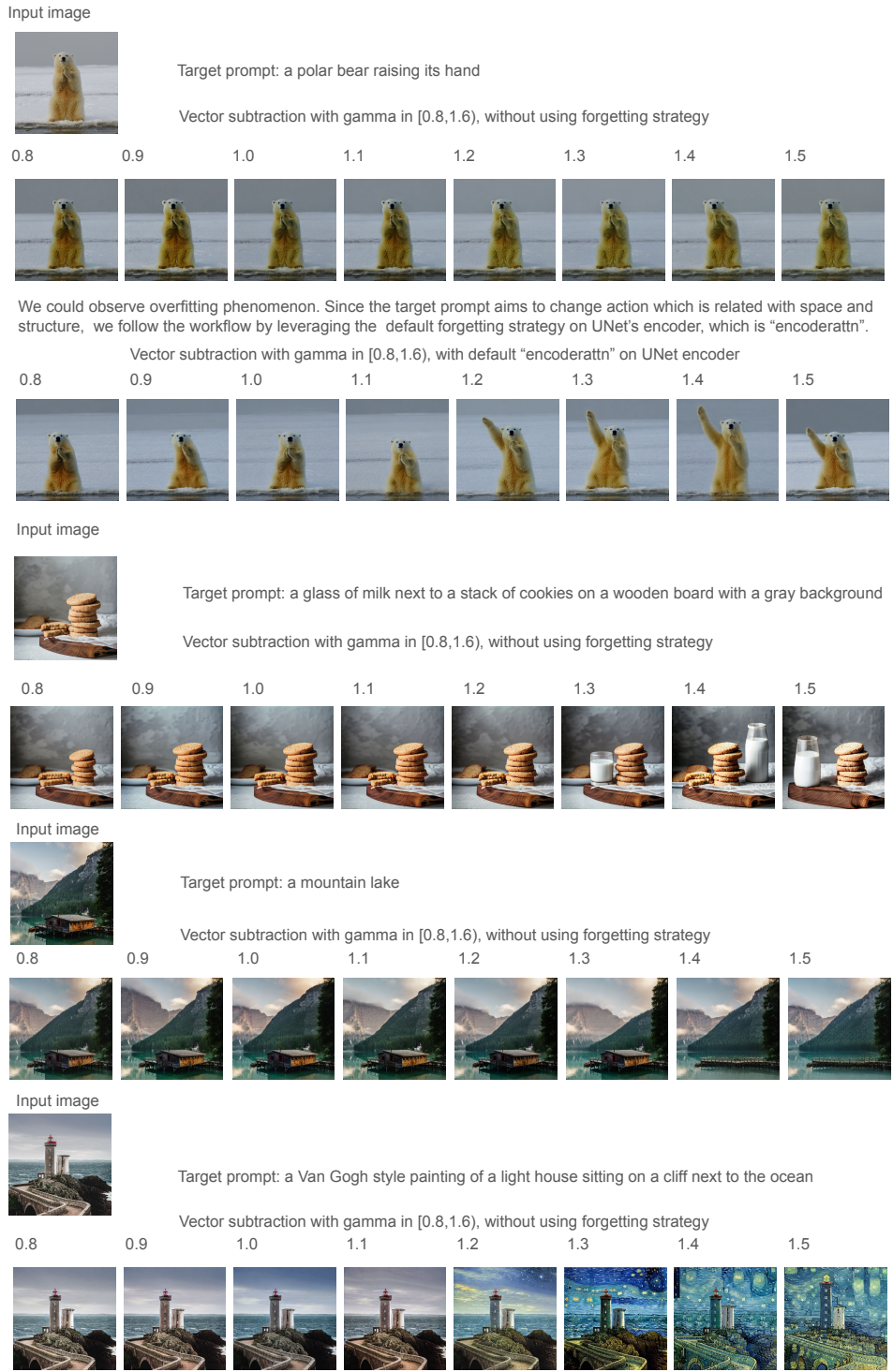}

   \caption{We show the practical workflow of Forgedit, with testing images from EditEval. In most cases, simple vector subtraction would finish the job. For other hard cases,  the default forgetting strategies, 'encoderattn' or 'decoderattn' according to editing intention on structrue or appearance, could solve the problems.  }
   \label{hyper}
\end{figure*}

 In this paper, we will demonstrate that the hyper-parameter tuning of Forgedit at editing stage is simple and show how to adjust these hyper-parameters with examples from  EditEval \citep{huang2024diffusion}.

The overall workflow is shown in Figure \ref{hyperworkflow}. No matter what the input image is, we use the same set of  hyper-paramters for finetuning stage. In the editing stage, the default workflow is to use vector subtraction with $\gamma$ in the range of 0.8 to 1.6. In general, a proper editing result should already been obtained from one of these 8 images. However, if a perfect editing did not show up, there are two possibilities, overfitting or underfitting. Underfitting leads to the fact that the edited object suffers from identity shift, which means with the editing strengthened, the appearance of target object becomes gradually inconsistent with input image. In this circumstance, one needs to apply vector projection instead, which I will show in another paper with examples from TEdBench. The more often case is overfitting, which means that Forgedit could reconstruct the input image well yet cannot conduct the edit successfully. With the disentangled property of UNet, we could utilize the forgetting strategy to tackle the overfitting issue.  If the target prompt aims to edit space and structure, one should use the default "encoderattn" forgetting strategy. If the target prompt aims to edit appearance and texture, one should use the default "decoderattn" forgetting strategy. Using  the examples from EditEvalv1 benchmark, we demonstrate several cases on how to adjust the hyper-parameters. The base model used in the following examples is Stable Diffusion 1.4.

For the first case where the input image is a polar bear on the ice field, the target prompt is "A polar bear raising its hand". To begin the workflow in Figure, we first run vector projection without forgetting strategy with with $\gamma$ in the range of 0.8 to 1.6. Shown in Figure , we could find that we are facing the overfitting issue and the polar bear is incapable of raising its hands. Following Figure , we then run the default forgetting strategy on UNet's encoder, i.e. "encoderattn", which means that newly learned parameters of self attention blocks and cross attention blocks are preserved in UNet encoder and all learned parameters of UNet decoder are preserved as well. The hyper-parameter  $\gamma$ still ranges from 0.8 to 1.6. This time we could find successful edits in the results.  For the complete command lines and hyer-paramters to reproduce our results in this paper, please check our official implementation of Forgedit \citep{Zhang_Pytorch_implementation_of_2024}.


\bibliography{iclr2024_conference}
\bibliographystyle{iclr2024_conference}

\appendix

\end{document}